# Action Prediction in Humans and Robots

F. Wörgötter[1,*], F. Ziaeetabar[1], S. Pfeiffer[1], O. Kaya[1], T. Kulvicius[1], M. Tamosiunaite[1,2],

[1]Universität Göttingen, Department for Computational Neuroscience at the Bernstein Center Göttingen, Inst. of Physics 3 and Leibniz Science Campus for Primate Cognition, Göttingen, Germany.

[2]Vytautas Magnus University, Faculty of Informatics, Kaunas, Lithuania.

## Abstract

Efficient action prediction is of central importance for the fluent workflow between humans and equally so for human-robot interaction. To achieve prediction, actions can be encoded by a series of events, where every event corresponds to a change in a (static or dynamic) relation between some of the objects in a scene. Manipulation actions and others can be uniquely encoded this way and only, on average, less than 60% of the time series has to pass until an action can be predicted. Using a virtual reality setup and testing ten different manipulation actions, here we show that in most cases humans predict actions at the same event as the algorithm. In addition, we perform an in-depth analysis about the temporal gain resulting from such predictions when chaining actions and show in some robotic experiments that the percentage gain for humans and robots is approximately equal. Thus, if robots use this algorithm then their prediction-moments will be compatible to those of their human interaction partners, which should much benefit natural human-robot collaboration.

## Summary

Event-based encoding of actions, earlier introduced for robotic action prediction, is similarly performed by humans, too.

---

[*] To whom correspondence should be sent at worgott@gwdg.de.

# Introduction

Experienced craftsmen will intuitively understand each other's actions even without words and their interaction is usually efficient and fluent. This is because cooperation between humans is heavily based on fast, mutual action prediction, originating already in non-human primates [1]. When we observe someone else, we can understand his or her actions before completion and even without any object present, like in the case of a pantomime. This allows us to perform our own actions early; seamlessly blending different action streams together similar to a dance. Currently existing artificial systems, on the other hand, are still not very good at action prediction. The fact that the same action can exist in so many variants, for example with different trajectories, objects, and object orientations (poses), makes machine prediction of action a hard problem.

Several methods have been employed to address this problem. Actions are most of the time analyzed using video data [2-4], frequently based also on the objects that take part in the action [5-8] and sometimes supported by an analysis of object affordances [9]. Alternatively, emphasis can also be laid on the spatio-temporal structure of the actions [3, 10] or on predictive models [11, 12]. Plan recognition is one other possible approach to analyze and predict actions [13-16]. Some approaches apply action grammars, which can be used to aid recognition and prediction [17-20]. Current deep learning based approaches implicitly use most of the aspects mentioned above [21-24]. Thus, the diversity and complexity of technical methods for action recognition and prediction is large and one can ask why it seems so effortless for humans to do this? For example, even a two-year old child will easily and quickly distinguish between her mother making a sandwich or doing the dishes. Children (and – of course – healthy adults) do not get lost in the details of the actual action execution. Somehow, we manage to extract the action's essence from which we quickly arrive at its semantics allowing us to predict it early.

During the last decade, others and we have tried to address action recognition from a more abstract, grammatical perspective. The essence of a sentence is in most languages given by the word order, where "Is there a cat?" is a question while "There is a cat." is a statement. This difference will prevail if we replace "cat" with "dog" and/or "there" with "this". Hence, the grammatical structure is, on its own (without considering specific words), a strong indicator of the basic meaning of a sentence. The analogy between actions and sentences is striking. Actions can fundamentally be distinguished by the sequence-order of the spatial and temporal relations between the different objects – including the hand – during the action [25-28]. This sequencing could be considered as the action's grammar [17, 20, 29, 30], where objects take the role of words.

Recently we introduced the framework of Extended Semantic Event Chains (ESEC), which is a highly compressed tabular action representation (*Figure 1, left*), encoding along its columns only the *temporal sequence of changes of object-object relations* in an action. The details of this encoding will be described below but first we would like to point out that the here shown ESEC of a "Hide" action, where one object is moved over another one to cover it, only contains seven changes (columns C1 to C7) that occur between its beginning and end. Using the ESEC encoding, it is possible to distinguish at least 35 different single-handed manipulation actions and we had argued that humans might not have many more available in their repertoire [31]. It is furthermore important to note that all these actions are recognized early. On average only 56% of the action duration needs to have passed before an action will be known ($P_E$ value of 43.94% in [31]). For the Hide action, this happens just a bit later, at column C4.

In the current study we fundamentally ask, whether humans – in the absence of object knowledge – would potentially use the same, underlying "algorithm" of analyzing object-object relations for action recognition and prediction. This is an important question as it specifically concerns the bootstrapping of action and object knowledge in very young children. The realm of (human made) objects is vast and it is a non-trivial cognitive problem for a child to understand their meaning. While 35 found manipulations might not capture all, this is, on the other hand, indicative of the fact that the number of actions is far more limited than that of objects. Would it not help if a child were able to understand actions without objects? For example, understanding the "essence of a cutting action" would allow bootstrapping object knowledge for various – never before seen – cutting tools. The fact that children, during their play, very often completely redefine and abuse things according to their action plan points to the primary role of action and lets these things become the action's objects, regardless of their adult meaning and common use (see OAC concept [32, 33]).

In the following, we present the results from a series of virtual reality-based experiments using moving colored blocks (*Figure 1, A-C*) instead of real objects asking our participants to indicate at what moment they would recognize the one action shown out of ten possible actions. The central novel contribution of this study is the finding that humans recognize the majority of these ten actions indeed at the same event column as the ESEC, pointing to the strong influence of object-object relations in action recognition. This may prove useful for seamless human-robot interaction and we also show how two different robots substantially gain performance speed when chaining actions in such a predictive manner.

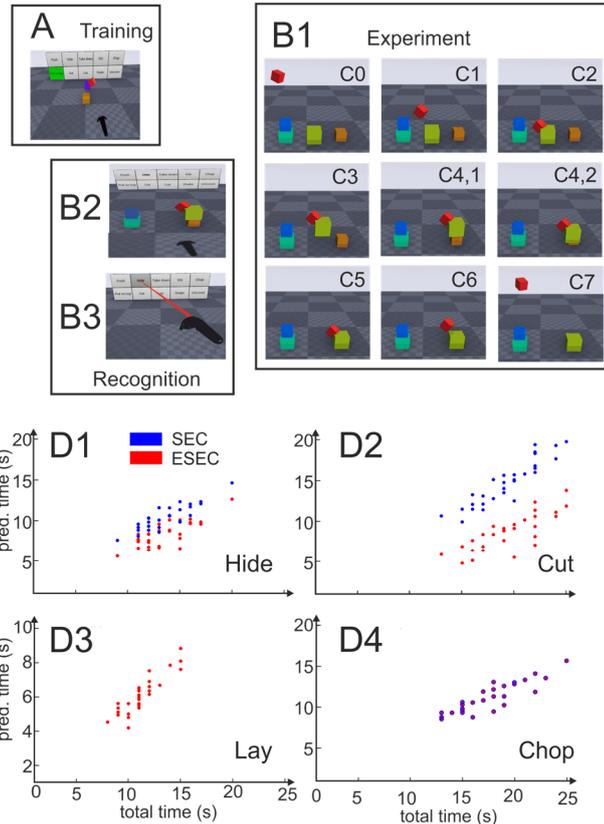

|     |      | C0 | C1 | C2  | C3 | C4  | C5 | C6 | C7 |
|-----|------|----|----|-----|----|-----|----|----|----|
| R1  | H, 1 | U  |    | T   |    |     |    | N  |    |
| R2  | H, 2 | U  |    |     |    | N   | A  |    |    |
| R3  | H, 3 | U  |    |     |    |     |    |    |    |
| R4  | H, G | N  |    |     |    |     |    |    |    |
| R5  | 1, 2 | U  |    |     |    | T   | A  |    |    |
| R6  | 1, 3 | U  |    |     |    |     |    |    |    |
| R7  | 1, G | U  |    | T   | N  |     | T  |    |    |
| R8  | 2, 3 | U  |    |     |    |     |    |    |    |
| R9  | 2, G | U  |    |     |    | T   | A  |    |    |
| R10 | 3, G | U  |    |     |    |     |    |    |    |
| R11 | H, 1 | U  |    | ArT |    |     |    | Ab | O  |
| R12 | H, 2 | U  |    |     |    | Ab  | A  |    |    |
| R13 | H, 3 | U  |    |     |    |     |    |    |    |
| R14 | H, G | O  | Ab |     |    |     |    | Ar | O  |
| R15 | 1, 2 | U  |    |     |    | To  | A  |    |    |
| R16 | 1, 3 | U  |    |     |    |     |    |    |    |
| R17 | 1, G | U  |    | To  | Ab |     | To |    |    |
| R18 | 2, 3 | U  |    |     |    |     |    |    |    |
| R19 | 2, G | U  |    |     |    | To  | A  |    |    |
| R20 | 3, G | U  |    |     |    |     |    |    |    |
| R21 | H, 1 | U  |    | HT  | MT |     | HT | MA | Q  |
| R22 | H, 2 | U  |    |     |    | GC  | A  |    |    |
| R23 | H, 3 | U  |    |     |    |     |    |    |    |
| R24 | H, G | Q  | S  |     | MA | GC  | S  |    | Q  |
| R25 | 1, 2 | U  |    |     |    | FMT | A  |    |    |
| R26 | 1, 3 | U  |    |     |    |     |    |    |    |
| R27 | 1, G | U  |    | HT  | MA | GC  | HT |    |    |
| R28 | 2, 3 | U  |    |     |    |     |    |    |    |
| R29 | 2, G | U  |    |     |    | HT  | A  |    |    |
| R30 | 3, G | U  |    |     |    |     |    |    |    |
|     |      | C0 | C1 | C2  | C3 | C4  | C5 | C6 | C7 |

*Figure 1: **Left**: Extended Semantic Event Chain (ESEC) of a "Hide" Action. **Right**: Experimental design **(A-C)** and relation between prediction and total time of four different actions **(D)**. Abbreviations in the ESEC are: T: touching, N: non-touching, A: absent, O: very far (static), Q: very far (dynamic), Ab: above, To: top, ArT: around with touch, S: stable, HT: halt together, MT: move together, FMT: fixed-moving together, MA: moving apart, GC: getting close. Note that the leftmost column, C0, is the same for all actions and indicates the start situation before any action.*

# Results

**ESEC encoding**

To be able to compare the VR results with those from the algorithmic relational analysis using ESECs we will first explain the ESEC encoding in some more detail. For ease of description, here we use object-words as for real actions but this description applies in the same way to the artificial cube-objects in the VR experiments.

Action start and end are well defined by the fact that the hand is at start *still* free (or at the end *again* free) not touching anything. When performing action recognition by our artificial systems, all relations are solely determined by vision similar to what a human observer would have to rely on. To arrive at a rigorous relational framework, object relations are determined from their bounding boxes (see Materials and Methods for details). This leads, for example, to the effect that the covering (Hide action) of an object by a cup is for the ESEC considered as a touching event when the cup is enclosing the object. In the VR experiments all events are of course *a priori* known, but to make this consistent with the above described "free-observation" of an action, the same detection method has been emulated here too ("simulated vision"). Following this, ESECs then encode actions in three sub-tables (*Figure 1, white, blue, orange colored rows, see left side*) using five types of objects: Hand, H; Ground, G; and *abstract* objects 1, 2, and 3. The latter are numbered by the occurrence of the corresponding *real* objects during an action. Object 1 is that object, which is first touched by the hand, Object 2 is the one first un-touched by Object 1, and Object 3 is the one first touched by Object 1. Hence, relations can exist between any of those five entities leading to ten rows in each sub-table. Note that in some actions, like the "Hide" action shown in *Figure 1*, not all objects play a role, resulting in empty (gray) rows in the table. Gray rows also occur if an object-object relation never changes during the action. Furthermore, note that objects "come into being" only at the moment when they are touched (or un-touched), before that they are "undefined" (U).

The white part of the ESEC encodes touching (T) and un-touching (N) events between these objects, the blue part their static relations (Ab: Hand is above Object 2, etc.), and the green part dynamic ones (GC: Hand and Ground are getting close, etc.). See Materials and Methods for the definition of the different relations, the geometrical assumptions to determine them and their thresholds.

**Setup and Procedures for the VR Experiments**

We have used a Vive VR headset and controller released by HTC in April 2016 which features a resolution of 1080 x 1200 per eye. The main advantage of that over competing headsets is its "room scale" system, which allows for precise 3D motion tracking between two infrared base stations. This provides the opportunity to record and review actions for the experiment on a larger scale of up to 5 meters diagonally. Thus, using human demonstration of each individual action we have implemented ten actions: Hide, Cut, Chop, Take down, Put on top,

Shake, Lay, Push, Uncover and Stir using differently colored blocks, where only the "Hand" was always red. Human demonstration results in jerk-free trajectories of the different moved blocks and actions look natural. For each action type, 30 different variants with different geometrical configurations and different numbers of distractors have been recorded (see Supplementary material for example VR-videos). We performed experiments with 49 participants (m/f ratio: 34/15, age range 20-68y, avg. 31.5y) showing them these 300 actions in random order. Before starting, we showed 10 actions to each participant for training, where the selection panel was highlighting the shown action in green (*Figure 1, A*). In the actual experiments – some frames for the Hide action are shown in panel B1 – subjects had to press a button on the controller at the moment when they believed that they had recognized the action (*Figure 1, B2*). After button-press, the scene disappeared to avoid post-hoc recognition and the subjects could, without time pressure, use a pointer (*Figure 1, B3*) to indicate, which action they recognized.

The frames shown in *Figure 1, B3* correspond to the columns in the ESEC on the left side. At C1 the Hand is above the ground and at C2 it touches Object 1 (yellow field). Through this several other relations come now also into being (other entries at C2). Object 1 leaves the ground at C3 and at C4 the yellow field shows that it is registered as "touching" Object 2 (see note on the simulated vision process above). This leads to the emergence of many static and dynamic relations (other entries in C4) and this is the column where the ESEC will unequivocally know that this is a Hide action. From there on, several more changes happen (C5-C7) until the action ends.

*Science Robotics* Page **7** of **27**

**Quantitative Analyzes**

Panels D in *Figure 1* plot the prediction times against the total times for four different VR-action types. Plots show results for ESECs (red) as well as SECs (blue), where in the latter case only the top sub-table (touching/non-touching) of the ESEC was used. This corresponds to the original SECs as introduced in our older studies [25]. In general, all relations are linear. Note that SECs cannot distinguish between several actions. For example, Lay is confused with Shake and therefore in panel D3 only ESEC performance is shown. For Chop (D4), ESECs and SECs are equally predictive. The linear relations seen here holds for all actions and this allows us to define a measure called "predictive power *P*", needed to analyze human results and to compare them to ESEC performance, by relating actual prediction time to total action time. The equation for *P* is given in the section on Materials and Methods (subsection: "Action prediction and quantification measure"). This measure essentially quantifies "how fast" someone will predict a certain action. For example, a *P*-value of 50 corresponds to action prediction happing after half of the action had been performed and 0 means that the action had to be completed before it was recognized.

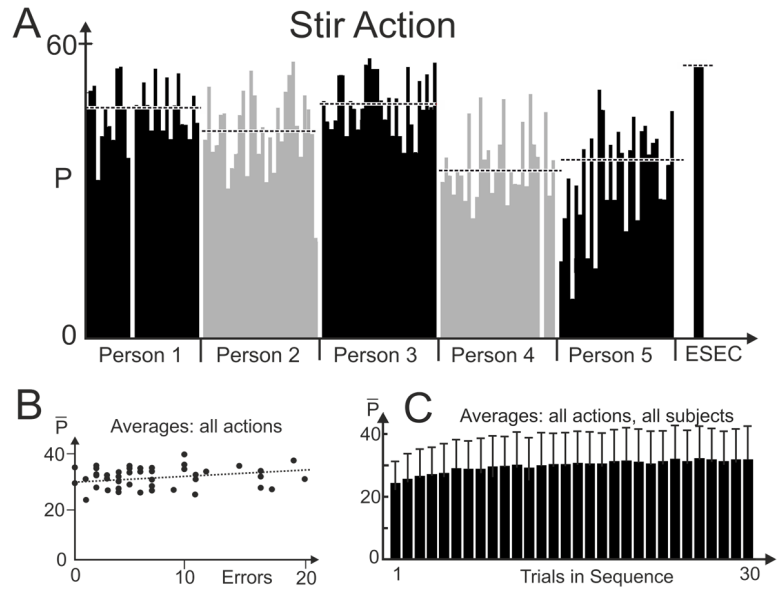

*Figure 2: (A) Individual performance, plotting predictive power P for all 30 trials of the Stir action of five subjects and the ESEC (median: dashed). (B) Predictive power averages $\bar{P}$ over all actions plotted against the number of recognition errors for all 300 trials for 49 subjects (dots). (C) Predictive power averages $\bar{P}$ over all actions and all subjects plotted against trials.*

*Figure 2A* shows values for *P* for five subjects demonstrating their performance for the Stir action. Some variability exists and white gaps are cases where no or wrong recognition has taken place. Subject 5 possibly shows small and gradual improvement over the first 10 trials indicative of a small degree of learning (see discussion of *Figure 2C*, below).

Subjects had *not* been prompted to specifically pay attention to speed. We avoided this to create a more realistic query situation potentially related to 'natural' action prediction in daily life (which usually happens also without the need of doing this speedily). Therefore, we observed only a small range of average predictive power values $\bar{P}$ across our 49 participants (*Figure 2B*). Interestingly, there is only a very small trend that faster participants, that have higher $\bar{P}$ values, produce more recognition errors than the slower ones.

We also asked to what degree learning takes place along the trials as suggested by the one case discussed above. The grand average $\bar{P}$ across all actions and subjects indeed shows a very small degree of increase for the first approximately five trials (*Figure 2C*), which is so small that we could ignore it for the remainder of this study.

*Figure 3A* shows how ESECs and humans differ in respect to predictive power for the ten different actions. In general, humans predict later than ESECs, where only for

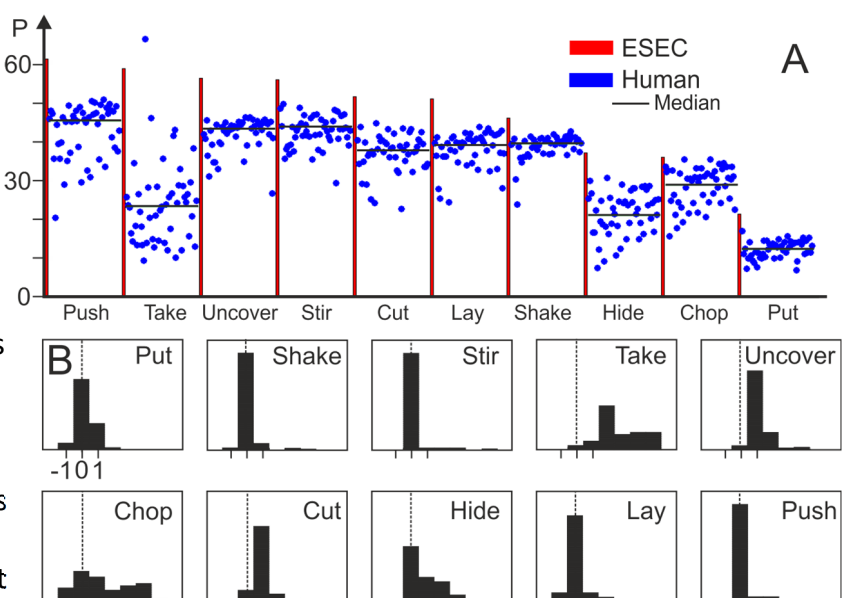

***Figure 3:*** *(A) Predictive power of ESECs (bars) and of 49 humans (dots) for all 30 trials of ten actions rank ordered according to ESEC performance. (B) Histograms of the frequency of human prediction in the same column than the ESEC (relative column number: 0, dashed line) or earlier/later (=left/right of dashed line). Total number (integral) in each histogram is about 1500 (exactly: 49x30 minus false recognitions).*

Take Down a strong difference exists. The most interesting question, however, is to ask at which event-column humans predict the actions. In *Figure 3B* we plot how often humans predict an action using the same event column (x-axis label: "0") as that where prediction happens in an ESEC, or earlier (x-axis: negative numbers) or later (positive numbers). With the exception of Chop and Take Down most distributions are narrow and – indeed – for six out of ten actions human prediction clearly happens most often at the same column as for the ESEC (for Chop the peak is also at column zero, but the distribution is too wide, preventing a clear conclusion, here). The fact that the predictive power of humans (*Figure 3A*) is lower than that of the ESECs is, very likely, owed to the fact that ESEC instantly predict when the relevant event happens, whereas humans probably take a bit more time to accumulate more evidence about the action.

**Robot Example Experiments and Human-based Performance Gain Quantification**

To demonstrate the temporal gain that good (versus less good) action prediction can achieve, we used two KUKA LWR robots equipped with Schunk 3-finger hands and let them "play a game". The target of this game was to perform as fast as possible five different actions on a set of hollow cubes; two by the one and three by the other robot. Robot 1 had at the beginning to perform one action, robot 2 had to recognize what robots 1 does and then perform another from the remaining four actions, and so on until all actions were done.

Before performing robotic tests, we analyzed human performance to select expressive cases for the experiments. For this, we could use the VR-experiments, because their actions came from human demonstration. From the existing ten actions, five were then selected to cover different situations: early versus late prediction times as well as large versus small differences between ESEC and SEC prediction times including one action, where SEC and ESEC prediction times do not differ. Following these requirements, we selected the actions: "put on top", "hide", "shake", "take down", and "push" for the robot experiments. Average timings, on which this selection was based, for these actions performed by humans are shown in *Table 1*.

|  | Average Action Duration [s] | Average ESEC Prediction Moment [s] | Average SEC Prediction Moment [s] |
|---|---|---|---|
| Take Down | 11.7 (2.9) | 3.3 (0.7) | 3.3 (0.7) |
| Put on Top | 12.0 (2.1) | 8.0 (1.9) | 9.2 (1.7) |
| Shake | 12.5 (2.1) | 6.5 (1.2) | 10.8 (1.7) |
| Push | 12.7 (1.9) | 5.0 (1.1) | 10.0 (1.6) |
| Hide | 13.8 (2.5) | 8.3 (1.6) | 10.3 (1.5) |

*Table 1:* *Human average performance measured from human demonstration used to create the VR experiments. Numbers in brackets are standard deviations.*

In the robotic experiments, we aim to compare action predictions using ESECs against those when using only SEC (touching/non-touching) information, where the baseline is the sequencing of those five actions without any predictive mechanism. The top part of *Figure 4* shows a timing diagram of one possible sequence "Hide-Shake-Take down-Push-Put" with timings (to scale) taken from five individual human VR actions and predictions based on ESECs. Agent 1 starts with the hiding action and at 8.5 s Agent 2 can predict this (leftmost downward arrow) and start its own shaking action. The leftmost red bar shows how many seconds have been saved by this prediction. In the end, this adds up to 25s of savings from a total non-interleaved chain length of 63s. This results in a ***P*** value of 39.7. The timing diagram also shows the complexity that any such 5-action chain may express. Often there are waiting gaps existing or an agent cannot yet make use of a prediction, because its own action has not yet finished (indicated by the slanted, dotted arrow).

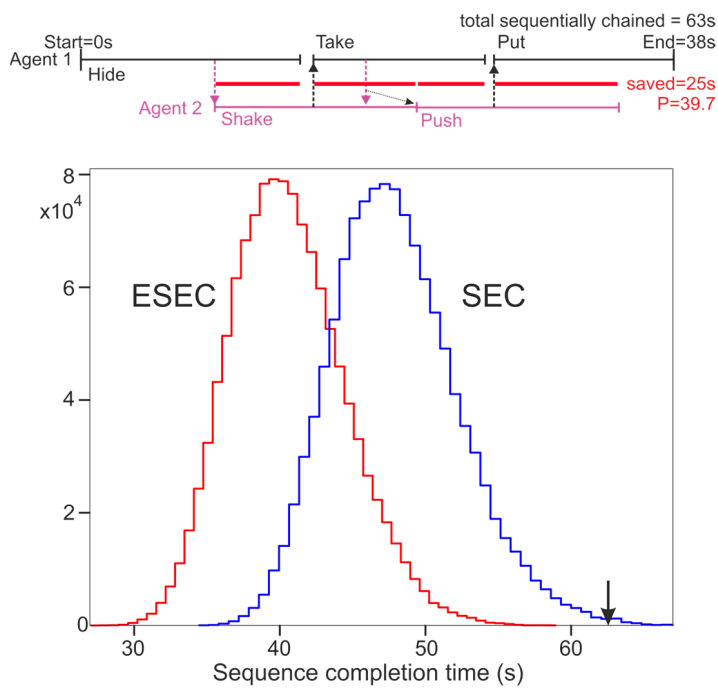

**Figure 4: Top:** Example of a sequence "Hide-Shake-Take down-Push-Put on top" using ESEC predictions and human action execution times (black and pink lines) for 2 actors. Dashed arrows show prediction moments and red lines the temporal savings. **Bottom**: Distribution of sequence completion times of humans based on 1.2 million combinations of the five actions as taken from the VR-actions setup. The arrow marks the average human sequence completion time of 62.6s.

In *Figure 4*, at the bottom, we show the distribution of human action sequence completion times that would arise when either using ESEC-based or SEC-based prediction for 1.2 million possible sequence combinations of the five different actions (for details see Materials and Methods) automatically creating and evaluating their timing diagrams. The total action sequence time without prediction is for humans on average 62.6 seconds (*Table 2*).

Note that humans (as shown above, *Figure 3*) seem to base their predictions on ESEC-related judgement. Thus, SEC-based sequencing does not occur in reality and we show the SEC-based distribution (*Figure 4*) and numbers (*Table 2*) only for comparison purposes.

The ESEC histogram demonstrates that action sequences would be completed within less than 30 up to more than 55 seconds where the actual numbers depend on the scene geometry. Also – as expected – the ESEC-histogram is centered at shorter times that the SEC-histogram, a fact that is also reflected by the averages in *Table 1*.

|   |   | Sequence Execution Time [s] and P values ||||||||
|---|---|---|---|---|---|---|---|---|---|
|   |   | HUMAN |||| ROBOT ||||
|   |   | Human total without prediction || Avg.: 62.6 (5.1) || Robot total without prediction || 1 | 75.2 |
|   |   |   |   |   |   |   |   | 2 | 83.1 |
|   |   |   |   |   |   |   |   | 3 | 75.9 |
|   | **Action Sequence** |   |   |   |   |   |   |   |   |
|   |   | Human ESEC | *P* | Human SEC | *P* | Robot ESEC | *P* | Robot SEC | *P* |
| 1 | **Take, Hide, Shake, Push, Put** | 37.8 (3.6) | *39.6* | 47.0 (3.7) | *24.9* | 44.3 | *41.1* | 52.7 | *29.9* |
| 2 | **Push, Put, Shake, Hide, Take** | 40.5 (3.9) | *35.3* | 51.9 (4.3) | *17.1* | 57.5 | *30.8* | 62.9 | *24.3* |
| 3 | **Put, Shake, Take, Hide, Push** | 42.1 (3.7) | *32.7* | 47.0 (3.6) | *24.9* | 47.0 | *38.1* | 50.9 | *22.9* |
|   | **Average P:** |   | ***35.9*** |   | *22.3* |   | ***36.7*** |   | *25.7* |

*Table 2: Performance of humans and robots for different sequencings of five actions and different prediction mechanisms (SEC vs. ESEC). The small font indicates that human – very likely – will not use a SEC-based prediction mechanism.*

For the robotic experiments, three example sequences (*Table 2, numbered 1 to 3*) of different sequencings, ordered by their human-ESEC-based completion times (from top to bottom: 1=fast, 2=medium, and 3=slow) have been investigated.

All robotic action parameters (trajectories and speed) had been predefined for the five different individual actions using dynamic movement primitives (DMPs, [34, 35]) to encode the motion from source to target. Hence, robots were not allowed to speed up or slow down their performance, because we are here interested in the predictive gain and not in any other performance characteristic. Timings for the robot execution times of any individual action had been taken from the average human completion times (*Table 1*). Still, some differences in action completion time will naturally arise due to different source-target distances. Furthermore, robots are about 15-20s slower than the humans (*Table 2, top*), for completing an action sequence, because of the quite slow speed of the robot hand for grasping and other hand-shape changes (see robot videos in the Supplementary Material).

*Table 2* compares average human performance (left part of the table) with robot performance on these three sequencings. Differences in sequencing times arise from the fact that some sequences allow for more efficient predictive chaining than others. A small font for human SEC performance is chosen in *Table 2* to indicate that this performance would never exist in reality. The right side of the table shows how fast the two KUKA robots can perform these sequences. As expected, ESEC-based execution is faster for humans as well as robots than SEC-based execution. Furthermore note that the ordering of 1,2,3=fast,medium,slow only holds for average human-ESEC-based execution times. Scene geometry as well as the fact that SECs have little predictive power lead to the situation that sequence completion times do not keep this ordering for all other cases. The median values of the ESEC predictive power of the here-chosen five actions range between $P=13.1$ and $P=46.0$ (*Figure 3*) with an average of $P=28.7$. The average predictive power values of the sequencings investigated in *Table 2* are with 35.9 (human) and 36.7 (robot) larger than that because time-savings from the predictions add up to some degree. Remarkably this gain is not terrific, because many cases exist where in an action chain an agent can actually not make much use of an early prediction, for example when it has not yet finished its own action (see e.g. the time line in *Figure 4, top*). Notably, predictive power values for human and robots are quite similar. The actual differences in individual execution times average out and the "gain" is for both types of agent comparable.

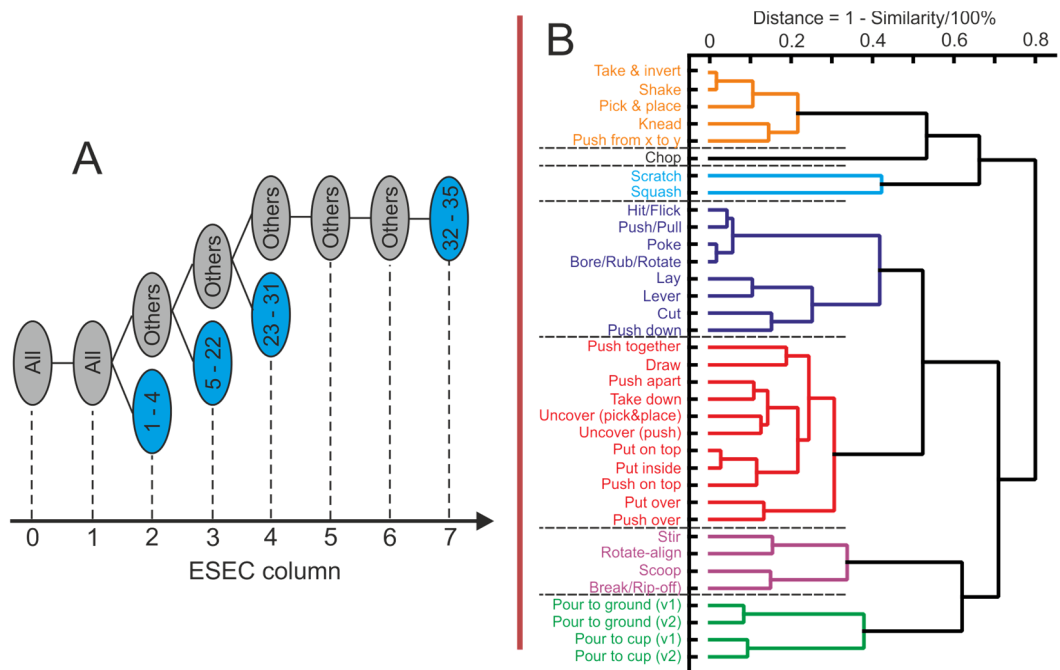

***Figure 5: (A)*** *Event columns where certain actions from an initial set of 35 actions are unequivocally recognized. For example, already at column 2, actions (1) Hit/Flick, (2) Poke, (3) Bore/Rub/Rotate, and (4) Lay will be known. The other actions are: (5) Push/Pull; (6) Stir; (7) Knead; (8) Lever; (9) Push from x to y; (10) Take & invert; (11) Shake; (12) Rotate-align; (13) Pick & place; (14) Pour from a container onto the ground when the liquid first un-touches the container then touches the ground (Pour to ground [v1]); (15) Pour from a container on the ground when the liquid can touch the container and the ground at the same time (Pour to ground [v2]); (16) Pour from a container to another container when the liquid first un-touches the container then touches another container (Pour to cup [v1]); (17) Pour from a container to another container when the liquid can touch the container and another container at the same time (Pour to cup [v2]); (18) Cut; (19) Chop; (20) Scratch; (21) Squash; (22) Draw; (23) Scoop; (24) Take down; (25) Push down; (26) Push apart; (27) Break/Rip-off; (28) Uncover by pick & place; (29) Uncover by push; (30) Put on top; (31) Put inside; (32) Push on top; (33) Push together; (34) Put over; (35) Push over.* ***(B)*** *Similarity Dendrogram clustering actions with a threshold of 0.5.*

**Discussion**

The framework of ESECs had been derived from older approaches that emphasize the "grammatical" structure of human actions [17, 20, 25-29]. ESECs carry substantial predictive power. *Figure 5 A* shows at which event column different actions can be predicted, when considering a total of 35 actions (Figure replotted from figure 9 in [31]). Column 7 is the latest moment and this corresponds, on average, to less than half of a complete ESEC. In the same older study, we had observed on different real data sets that the average predictive power (***P****-*value) for ESECs is above 60 as compared to a standard Hidden Markov Model-based prediction with P<35. Furthermore, when calculating a similarity tree diagram (*Figure 5 B*, replotted from figure 6 in [31]) one can see that ESECs group actions roughly in the same way as humans would (with few exceptions). Hence, it appears that ESECs represent essentially a human-like action semantics. This had prompted us to as in the current study whether humans and ESECs "preform in the same way" when predicting actions. The central finding of this study is that for most of the tested actions, indeed humans and ESECs use the same event configuration (column) for recognition. Humans take a little longer, though (blue dots in *Figure 3* are lower than the ESEC-bars). This is possibly because humans accumulate a bit more evidence until they make a prediction than ESEC. For example, a cutting action will be predicted by the ESEC at the very first millisecond where the back-and-forth cutting movement starts; humans take a bit longer until they can be sure that one cube really emulates a knife and performs cutting.

Another important aspect of the ESEC framework is that it neglects timing. At first, this appears problematic, because one loses finer distinctions between dynamic actions. Here we would, however, argue that most often action dynamics play a clear role in distinguishing between skilled versus less-skilled performance (e.g. a tennis expert versus a beginner), while dynamics are less relevant for action recognition and prediction per se.

On the contrary, losing time-information is a feature that helps grouping actions together in the right way. The individual action timings *Table 1* show substantial standard deviations and the ESEC-based prediction distribution (red curve in *Figure 4*) shows that action-sequence timings can vary widely, too. Thus, time may rather confuse than help recognition and prediction. Furthermore, in our example robotic experiments we found that the ESEC-based predictive power values for humans and robots accumulated over five actions differ only by 0.8 (human: *P*=35.9, robot: *P*=36.7) in spite of the fact that the robots performed the sequence much slower than humans would have. This finding is interesting, too. If robots use the events in the ESECs to predict actions then their interaction with a human should be smooth and "cognitively seamless" in spite of different action speeds , because – under these experimental

conditions – humans appear to use the same (or temporally close) events for their own action prediction.

Taken together we would argue that a grammatical view onto actions may underlie human action understanding and that this type of encoding should be helpful for robots especially when having to operate together with us.

## Materials and Methods

The core of our work relies on the Enriched Semantic Event Chain (ESEC) framework. The main concept had been described above and here we summarize it briefly and add all missing details. An ESEC is a table consisting of three sets of 10 rows each (Figure 1). The first set encodes the changes of the touching (T) and non-touching (N) relations for each pair of objects during a manipulation. The second set captures static spatial relations (SSR) and the third set dynamic spatial relations (DSR) between the objects. Hence, a new column is created whenever a change in any of these relations occurs. Objects are labeled as H, 1, 2, 3, and G, where H and G are "Hand" and "Ground" and Objects 1, 2, and 3 are numbered by their occurrence during the action as explained in the main text above. Actual object names are not relevant for ESEC encoding, because this encoding relies only on the sequentiallity of the different object-object relation changes. Defining ESECs with this kind of strict order of object-appearance leads to always the same row ordering in the ESEC matrix.

In the following we will describe, how the manipulated objects are modeled and – according to these models – how static and dynamic spatial object-object relations are defined.

### Object Relations

**Touching/non-touching and other general relations:** We interpret touching/non-touching (T versus N) as the occurrence or non-occurrence of collisions between point clouds based on the k-d tree algorithm. Moreover, there are some other possible types of simple relations between objects in our framework.

- Undefined or "U" when an object does not exist.
- Destroyed or "X" when an object is destroyed or loses its primary shape (e.g. in cut, chop, scoop, or break actions).
- Absent or "A" when an object has already been there, but has now disappeared (e.g. in a hide action).

**Static and dynamic object relations**

Point cloud collisions using the k-d tree algorithm will not suffice to define static and/or dynamic object relations. For this, a different treatment of the objects is needed.

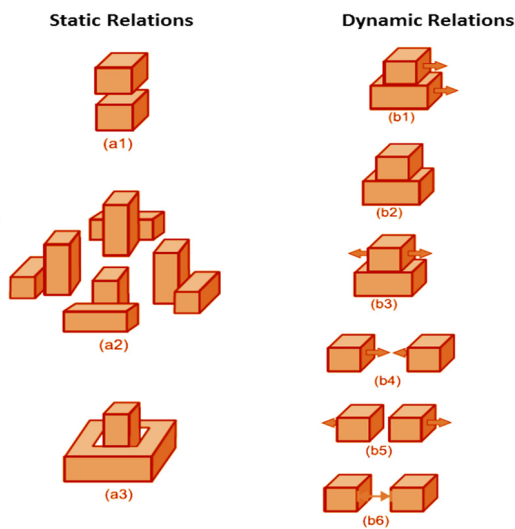

*Figure 6:* (a) Static Spatial Relations: (a1) Above/Below, (a2) Around, (a3) Inside/Surround. (b) Dynamic Spatial Relations: (b1) Moving Together, (b2) Halting Together, (b3) Fixed-Moving Together, (b4) Getting Close, (b5) Moving Apart, (b6) Stable.

**Object modeling for static and dynamic object relations**

Hence, we are approximating each object by an Axis Aligned Bounding Box (AABB). In this model all coordinate axes are aligned according to the direction of the camera axes. Note that the camera is fixed and does not move during the manipulations. All relations have been then defined relative to this setting.

The x axis corresponds to the right/left relation, while y and z axes define the directions of the above/below and the front/back relations, respectively.

An AABB models a point cloud by a cube with sides parallel to the directions of the coordinate system axes. AABB computation details are discussed in [31].

***Static Spatial Relations*** depend on the relative position of two objects in space. These types of relations are computed at every frame and there is no need for data from previous frames. We define the following types of SSRs: ''Above'' (**Ab**), ''Below'' (**Be**), ''Right'' (**R**), ''Left'' (**L**), ''Front'' (**F**), ''Back'' (**Ba**), ''Inside'' (**In**), ''Surround'' (**Sa**) and ''Between'' (**Bw**). Right, Left, Front and Back are composed into "Around" (**AR**) and used at times when one object is surrounded by the other. Moreover, "Above", "Below" and "Around" relations can be combined with the "touching" relation and are then converted to "Top" (**To**), "Bottom" (**Bo**) and "Around with touch" (**ArT**), respectively. *Figure 6 (a1-a3)* represents static relations between two objects in term of cubes. Note that this way this type of encoding contains a bit of redundancy for better human readability of these tables.

If the distance between two objects' AABBs exceeds a certain threshold, they do not have any of the above mentioned relations and their static relation is assumed as Null (**O**). Therefore, the set of static spatial relations is given by: SSR = {Ab, Be, R, L, F, Ba, Ar, To, Bo, ArT, In, Sa, Bw, O}.

Each relation is defined by a set of rules.

In general, xmin, xmax, ymin, ymax, zmin and zmax are the minimum and maximum values between the points of the object Θ's AABB of at the $i_{th}$ frame along the x, y and z-axes, respectively.

Let us consider the relation "Above": SSR($Θ_i$, $Θ_j$) = Ab (object $Θ_i$ is above the object $Θ_j$) if ymin($Θ_i$) < ymin($Θ_j$) and ymax($Θ_i$) < ymax($Θ_j$) as well as all the following (exception) conditions are not true: xmin($Θ_i$) > xmax($Θ_j$) and xmax($Θ_i$) < xmin($Θ_j$); zmin($Θ_i$) > zmax($Θ_j$) and zmax($Θ_i$) < zmin($Θ_j$);

The exception conditions exclude from the relation "Above" those cases when two AABBs do not overlap in right/left (x direction) or front/back (z direction). Several examples of objects holding relation SSR(red, blue) = Ab are shown in *Figure 7*, when the size and shift in x direction varies.

SSR($Θ_i$, $Θ_j$) = Be (below) is defined by ymin($Θ_i$) > ymin($Θ_j$) and ymax($Θ_i$) > ymax($Θ_j$) and the same set of exception conditions. The relations R, L, F, Ba are defined in a similar way. For R and L the emphasis is on the "x" dimension, while for the F, Ba the emphasis is on the "z" dimension. For the relation "inside", SSR($Θ_i$, $Θ_j$) = In, x and z coordinates of AABB $Θ_i$ must be between the x and z coordinates of AABB $\Theta_j$ respectively while $ymin(\Theta_j) < ymax(\Theta_i) \leq ymax(\Theta_j)$. Surround (Sa) relation needs opposite conditions.

For the relation "in between" (Bw), we first define the "between space" for two objects, which is given by extending the AABBs from two non-interacting objects toward each other along the camera axis and considering the intersection of these extensions. Whenever the AABB of a third object remains completely in the "between space" of the two other objects, it is assumed that the third object is "in between" (Bw) of them. The rules relating to this relation are defined by SSR ($\Theta_i$, $\Theta_k$, $\Theta_j$) = Bw (object $\Theta_k$ is between the objects $\Theta_i$ and $\Theta_j$): $xmin(\Theta_k) \geq mini(xmax(\Theta_i), xmax(\Theta_j))$ and $xmax(\Theta_k) \leq max(xmin(\Theta_i), xmin(\Theta_j))$ and $ymin(\Theta_k) \geq max(ymin(\Theta_i), xmin(\Theta_j))$ and $ymax(\Theta_k) \leq min(ymax(\Theta_i), xmax(\Theta_j))$ and $zmin(\Theta_k) \geq max(zmin(\Theta_i), xmin(\Theta_j))$ and $zmax(\Theta_k) \leq min(zmax(\Theta_i), xmax(\Theta_j))$.

There could be more than one static spatial relation between two objects, e.g. one object's AABB can be both to the left and in the back of the other

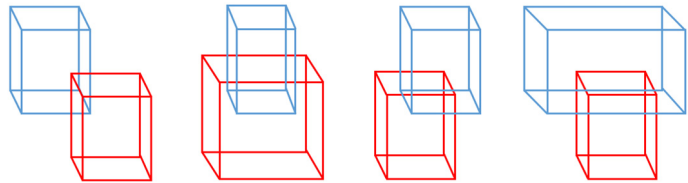

*Figure 7:* Possible states of Above-Below relations between two AABBs when size and x positions vary.

object's AABB. However, we disallow this and define only one relation per object pair.

To achieve this the following procedure is adopted. Each AABB is a cube, which includes six rectangular surfaces. We label them as top, bottom, right, left, front and back based on their positions in our scene coordinate system. When object $\Theta_i$ is to the left of object $\Theta_j$, one can make a projection from the right surface of object $\Theta_i$ onto the left rectangle of object $\Theta_j$ and consider only the rectangle intersection area. We call this area the "shadow".

Suppose SSR($\Theta_i$, $\Theta_j$) = {$R_1$, . . . , $R_k$} whith $R_1$, . . . , $R_k \in$ SSR. We calculate the *shadow*($\Theta_i$, $\Theta_j$, $R_m$) (1 $\leq m \leq k$) for all relations $R_m$ between objects $\Theta_i$ and $\Theta_j$. Then we select the relation with the biggest shadow as the main static relation between the two objects, SSR($\Theta_i$, $\Theta_j$) = $R_n$(1 $\leq n \leq k$), if: shadow($\Theta_i$, $\Theta_j$, $R_n$) = $max_{1 \leq m \leq k}$ (Shadow($\Theta_i$, $\Theta_j$, $R_n$)).

Several of these static relations are dependent on the viewpoint and the exact relation is often not relevant (also humans do not consider this many times). For instance, when picking up a spoon to stir a cup of tea, it is usually not important that the spoon is picked up from the right or the left side of the cup. Thus, we define in addition a relation called "Around" (Ar) and gather all relations L, R, F, Ba into it. This way, "Ar" (Around) contains the space located *lateral* to the object in a limited radius equal to a threshold π. This space does not cover vertical neighborhoods like "Above" or "Below" [31].

**Dynamic Spatial Relations** define the spatial relation between two objects that can be in a moving or stable condition. Here, different from SSRs, which are computable at every frame, we need some information from the previous *N* frames (e.g., distance related parameters) between each pair of objects. The parameter *N* is defined according to the frame-rate of the movie, where we determine *N* as frame count for covering 0.5 s. This has turned out to be a good heuristic estimate of the time it takes for a person to change the relations between objects. Therefore, if the video rate is μ frames per second, then $N = 0.5\mu$. DSRs include the following relations: "Moving Together" (**MT**), "Halting Together" (**HT**), "Fixed-Moving Together" (**FMT**), "Getting Close" (**GC**), "Moving Apart" (**MA**) and "Stable" (**S**). These dynamic spatial relations between two objects are shown in *Figure 6 (b1–b6)* by using cubes. MT, HT and FMT define situations when two objects are touching each other while: both of them are moving in a same way (MT), are not moving (HT), or when one object is fixed and does not move, while the other one is moving on or across it (FMT). Case S (stable) denotes that any distance-change between objects is less than a defined threshold (here, we have considered this threshold as $\varepsilon = 1$ cm) and remains constant during the action sequence. In addition, Q is used to indicate a dynamic relation between two objects if the distance between them is more than the defined threshold $\mho = 10$ cm or if they do not have any of the above defined dynamic relations. Therefore, we have defined DSR as a two argument function where the arguments are AABB cubes in the scene. Suppose $c_i^f$ shows the central point of the AABB of object $e_i^f$ (object $e_i$ in $f_{th}$ frame). For measuring the Euclidean distance between AABBs of $e_i$ and $e_j$ in the $f_{th}$ frame, $\delta(e_i^{f+\partial}, e_j^{f+\partial}) = ||c_i^f - c_j^f||$ is defined by:

$$DSR(e_i^f, e_j^f) = \begin{cases} GC & if\ \delta(e_i^{f+\partial}, e_j^{f+\partial}) - \delta(e_i^f, e_j^{f+\partial}) < \xi \\ MA & if\ \delta(e_i^{f+\partial}, e_j^{f+\partial}) - \delta(e_i^f, e_j^{f+\partial}) > \xi \end{cases}$$

We use a time window $\partial$=10 frames in our experiments (recording speed is 30 fps); the threshold $\xi$ is kept at 10 cm.

When calculating dynamic relations, we are also checking the touching relations between those two objects. For this we first define TNR, as a two argument function that illustrates whether two objects are touching or not-touching each other. This function is then used below to define several conditions:

**Con1**: $TNR(e_i^f, e_j^f) = T$ && $TNR(e_i^{f+\partial}, e_j^{f+\partial}) = T$

**Con2**: $TNR(e_i^f, e_j^f) = N$ && $TNR(e_i^{f+\partial}, e_j^{f+\partial}) = N$

**Con3**: $c_i^f \neq c_i^{f+\partial}$

**Con4**: $c_j^f \neq c_j^{f+\partial}$

**Con5**: $\delta(e_i^{f+\partial}, e_j^{f+\partial}) - \delta(e_i^f, e_j^{f+\partial}) < \xi$

Now, the dynamic relations MT, HT, FMT and S are defined based on the above condition in the following way:

$$DSR(e_i^f, e_j^f) = \begin{cases} MT, & if\ Con1\ \&\&\ Con3\ \&\&\ Con4 \\ HT, & if\ Con1\ \&\&\ \sim Con3\ \&\&\ \sim Con4 \\ FMT, & if\ Con1\ \&\&\ (Con3\ XOR\ Con4) \\ S, & if\ Con2\ \&\&\ Con5 \end{cases}$$

**Similarity Measure**

Note that all actions shown in the VR experiments had been generated by capturing their movement trajectories from human demonstration. In addition, different geometrical arrangements of objects and distractors were used. As a consequence, ESECs from the same action class are normally quite similar but often not identical. (Obviously, this statement is also true for ESECs that are extracted from observing human actions directly.)

Thus, we need a method to calculate the similarity of two manipulation actions. In general, different measures would be possible and the one used here (described below) had been chosen, because it discriminates all 35 actions, while keeping their sematic similarities ($Figure\ 5,\ B$) in a way similar to human general judgment.

Different from SECs, where only one type of relation exist, here we have to deal with three types, which can occur concurrently: touching/non-touching, SSR and DSR.

We proceed as follows: Suppose α1 and α2 are two actions. Their ESEC matrices have n and m columns, respectively. Instead of considering their 30-row ESECs individually we concatenate the corresponding T/N, SSR and DSR of each fundamental object pair into a vector with three components and now create a 10-row matrix for α1 and α2. We annotate α1 with components x and α2 with components y. Thus, α1 becomes:

$$\alpha 1 = \begin{pmatrix} (x_{1,1}, x_{11,1}, x_{21,1}) & (x_{1,2}, x_{11,2}, x_{21,2}) & \cdots & (x_{1,n}, x_{11,n}, x_{21,n}) \\ (x_{2,1}, x_{12,1}, x_{22,1}) & (x_{2,2}, x_{12,2}, x_{22,2}) & \cdots & (x_{2,n}, x_{12,n}, x_{21,n}) \\ & & \cdots & \\ (x_{10,1}, x_{20,1}, x_{30,1}) & (x_{10,2}, x_{20,2}, x_{30,2}) & \cdots & (x_{10,n}, x_{20,n}, x_{30,n}) \end{pmatrix}$$

With these components of both matrices we define the differences in the three different relationship categories $D^{1:3}$ by:

$$D^1_{i,j} = \begin{cases} 0, & if\ x_{i,j} = y_{i,j} \\ 1, & otherwise \end{cases}$$

$$D^2_{i,j} = \begin{cases} 0, & if\ x_{i+10,j} = y_{i+10,j} \\ 1, & otherwise \end{cases}$$

$$D^3_{i,j} = \begin{cases} 0, & if\ x_{i+20,j} = y_{i+20,j} \\ 1, & otherwise \end{cases}$$

where 1 ≤ i ≤ 10, 1 ≤ j ≤ P, P = max(n, m)

Then we define the composite difference for the three categories as follows:

$$diff_{i,j} = \sqrt{D^1_{i,j} + D^2_{i,j} + D^3_{i,j}}$$

If one matrix has more columns than the other matrix, i.e. m <n or vice versa, we repeat the last column of the smaller matrix until getting the same number of columns as the bigger matrix. Now *Difference* is defined as a matrix, where its components represent the difference values between the two ESEC s' corresponding components.

$$Difference_{(10,p)} = \begin{pmatrix} diff_{1,1} & diff_{1,2} & \ldots & diff_{1,p} \\ diff_{2,1} & diff_{2,2} & \ldots & diff_{2,p} \\ & & \ldots & \\ diff_{10,1} & diff_{10,2} & \ldots & diff_{10,p} \end{pmatrix}$$

Where $diff_{i,j}$ denotes the dissimilarity of the $i_{th}$ objects pair at the $j_{th}$ time stamp (column). Then, *Dis*, which is the total dissimilarity between ESECs of α1 and α2 is obtained as the average across all components of the matrix *Difference*.

$$Dis_{\alpha 1, \alpha 2} = \frac{1}{p*10} \left( \sum_{j=1}^{p} \sum_{i=1}^{10} diff_{i,j} \right)$$

Accordingly, $Sim_{\alpha 1, \alpha 2}$, the similarity between the ESECs matrices of actions α1 and α2, is obtained as the complimentary value of *Dis*.

$$Sim_{\alpha 1, \alpha 2} = (1 - Dis_{\alpha 1, \alpha 2}) * 100$$

This is, thus, given in percent and used to quantify the similarity between any two actions in this study.

**Action prediction and quantification measure**

The variability of the ESECs within any given action class requires using a probabilistic method to assess at what time point any action would be predictable by its ESEC. For this we causally compare – as the columns arrive one after the other – that action's ESEC to the same number of columns of 20 each randomly chosen ESECs from all action classes. This is done by calculating for this (growing) ESEC the similarity values to every member in the comparison-set. We label an action as "predicted" when the average similarity is still high for one class whereas it has dropped to a low value for all other classes.

The moment at which the prediction happens is called *prediction moment* and is shown as *T(α)*. Based on that, prediction power *P* is defined as:

$$P = \left(1 - \frac{T(\alpha)}{Tot(\alpha)}\right) * 100\%,$$

where *Tot(α)*, is the total time of the action video. The moment where the hand appears in the scene and leaves the scene are considered as the initial and last moment, respectively.

**Statistical Evaluation of Human Action Chaining**

The distributions in *Figure 4* were calculated by randomly choosing one out of the 30 existing realizations for each of the five actions and repeating this 10000 times, this way creating a base set. For each sample in the base set, all possible 120 permutations were then analyzed creating a total of 1,200,000 cases. For every case, the sequence completion (execution) duration was calculated when performing action prediction either using SECs or ESECs and added to the histogram. All VR setups represent realistic geometrical configurations. Hence, the resulting, histograms from more than 1 million piece-wise combined actions appear trustworthy for human action sequencing of the here-chosen five actions.

**Robotic Experiments, Experimental setup**

Our robotic setup consists of two KUKA LWR robotic arms with Schunk three-finger hands (SDH2) and an ASUS-Xtion RGB-D sensor. The scene consisted of three smaller and four bigger wooden boxes, where two different initial configurations were used (*Figure 8*). Bottom parts of the bigger boxes were removed to make hiding actions possible. The following three action sequences were performed: 1) Take down, hide, shake, push, put on top; 2) Push, put on top, shake, hide, take down; and 3) Put on top, shake, take down, hide, push (the video is available at the Supplementary Material). Each of these action sequences were performed by using both frameworks, i.e., SEC and ESEC (in total six experiments).

**Robotic Experiments, Action execution**

Five main actions (hide, take down, put on top, push, shake) and two supportive actions (approach and leave an object) were recorded using kinesthetic guidance and making use of our library-of-action encoding [36, 37] by which they were encoded using dynamic movement primitives (DMPs, [34, 35]). Note, that durations of motions were scaled in proportion to the path length to have constant velocity on average for all motions. All visual analyses were performed using our previously developed methods [38, 39].

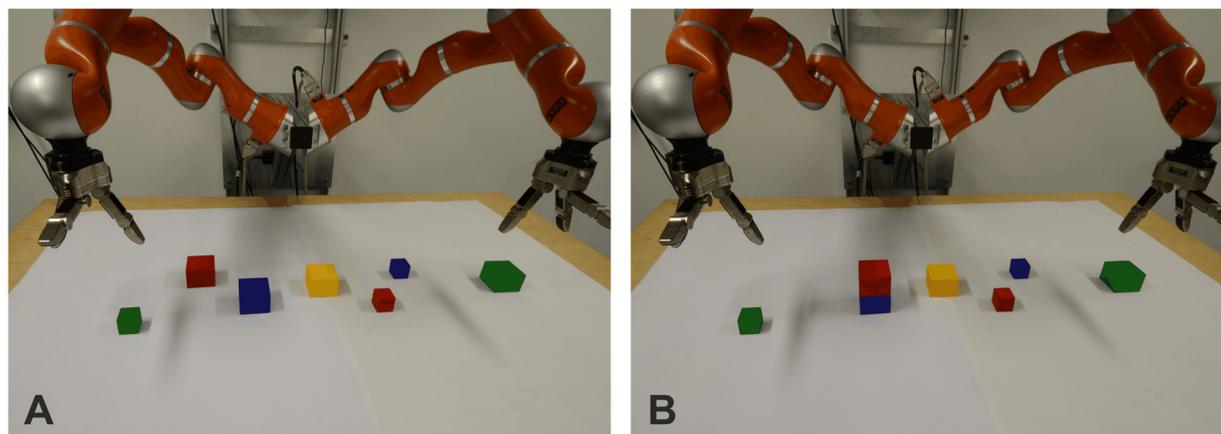

*Figure 8:* Two initial configurations used for robot experiments. The configuration shown on the left side was used for the action sequence 1 (take down, hide, shake, push, put on top), whereas configuration shown on the right side was used for the action sequence 2 (push, put on top, shake, hide, take down) and action sequence 3 (put on top, shake, take down, hide, push).

**Acknowledgments**

The authors are especially grateful to Ricarda Schubotz and Nadiya El-Sourani for their help at various stages in the design of the experiments and the interpretation of the results.

**Funding:** The research leading to these results has received funding from the German Research Foundation (DFG) grant WO388/13-1 and the European Community's H2020 Programme (Future and Emerging Technologies, FET) under grant agreement no. 732266, Plan4Act.

**Author contributions:** F.W. and M.T. designed the experiments and guided the analyzes, where F.W. wrote the paper, F.Z. performed the VR experiments and corresponding data analysis, S.P. programmed the VR environment, O.K. and T.K. performed the robot experiments.

**Competing interests:** There are no competing interests.